\documentclass[10pt,twocolumn,letterpaper]{article}

\usepackage{iccv}              
\usepackage{textcomp}
\usepackage{amsmath}
\usepackage{multirow}
\usepackage{colortbl}
\usepackage{pifont}
\definecolor{iccvblue}{rgb}{0.21,0.49,0.74}
\definecolor{mblue}{rgb}{0.00, 0.44, 0.75}
\definecolor{mred}{rgb}{0.75, 0.00, 0.00}
\usepackage[pagebackref,breaklinks,colorlinks,allcolors=iccvblue]{hyperref}

\def\paperID{4712} 
\def\confName{ICCV}
\def\confYear{2025}

\title{ReferDINO: Referring Video Object Segmentation with Visual Grounding Foundations}

\author{
\textbf{Tianming Liang\textsuperscript{1}  \ \ 
Kun-Yu Lin\textsuperscript{1} \ \ 
Chaolei Tan\textsuperscript{1} \ \ 
Jianguo Zhang\textsuperscript{2} \ \ 
Wei-Shi Zheng\textsuperscript{1} \ \ 
Jian-Fang Hu\textsuperscript{1}\thanks{Corresponding author.}}\\
\textsuperscript{1}Sun Yat-sen University \ \ 
\textsuperscript{2}Southern University of Science and Technology\\
{\tt\small liangtm@mail2.sysu.edu.cn, 
hujf5@mail.sysu.edu.cn
}\\\vspace{-.5em}\\
Project page: \url{https://isee-laboratory.github.io/ReferDINO}
}

\begin{document}
\maketitle
\begin{abstract}
Referring video object segmentation (RVOS) aims to segment target objects throughout a video based on a text description. 
This is challenging as it involves deep vision-language understanding, pixel-level dense prediction and spatiotemporal reasoning.
Despite notable progress in recent years, existing methods still exhibit a noticeable gap when considering all these aspects.
In this work, we propose \textbf{ReferDINO}, a strong RVOS model that inherits region-level vision-language alignment from foundational visual grounding models, and is further endowed with pixel-level dense perception and cross-modal spatiotemporal reasoning. 
In detail, ReferDINO integrates two key components: 
1) a grounding-guided deformable mask decoder that utilizes location prediction to progressively guide mask prediction through differentiable deformation mechanisms;
2) an object-consistent temporal enhancer that injects pretrained time-varying text features into inter-frame interaction to capture object-aware dynamic changes.
Moreover, a confidence-aware query pruning strategy is designed to accelerate object decoding without compromising model performance.
Extensive experimental results on five benchmarks demonstrate that our ReferDINO significantly outperforms previous methods (e.g., +3.9\% \(\mathcal{J}\&\mathcal{F}\) on Ref-YouTube-VOS) with real-time inference speed (51 FPS).
\end{abstract}

\section{Introduction}
\label{sec:intro}

Referring Video Object Segmentation (RVOS)~\cite{botach2022end,wu2022language} aims to segment the target object in a video, referred by a given text description. 
This emerging task is potentially benefitial for many interactive video applications, and has therefore attracted great attention in the computer vision community.
Compared to the unimodal video segmentation tasks~\cite{perazzi2016benchmark,xu2018youtube}, RVOS is more challenging since it requires a strong vision-language capability to understand complicated descriptions and associate visual objects with text.

\begin{figure}[t]
    \centering
    \includegraphics[scale=0.42]{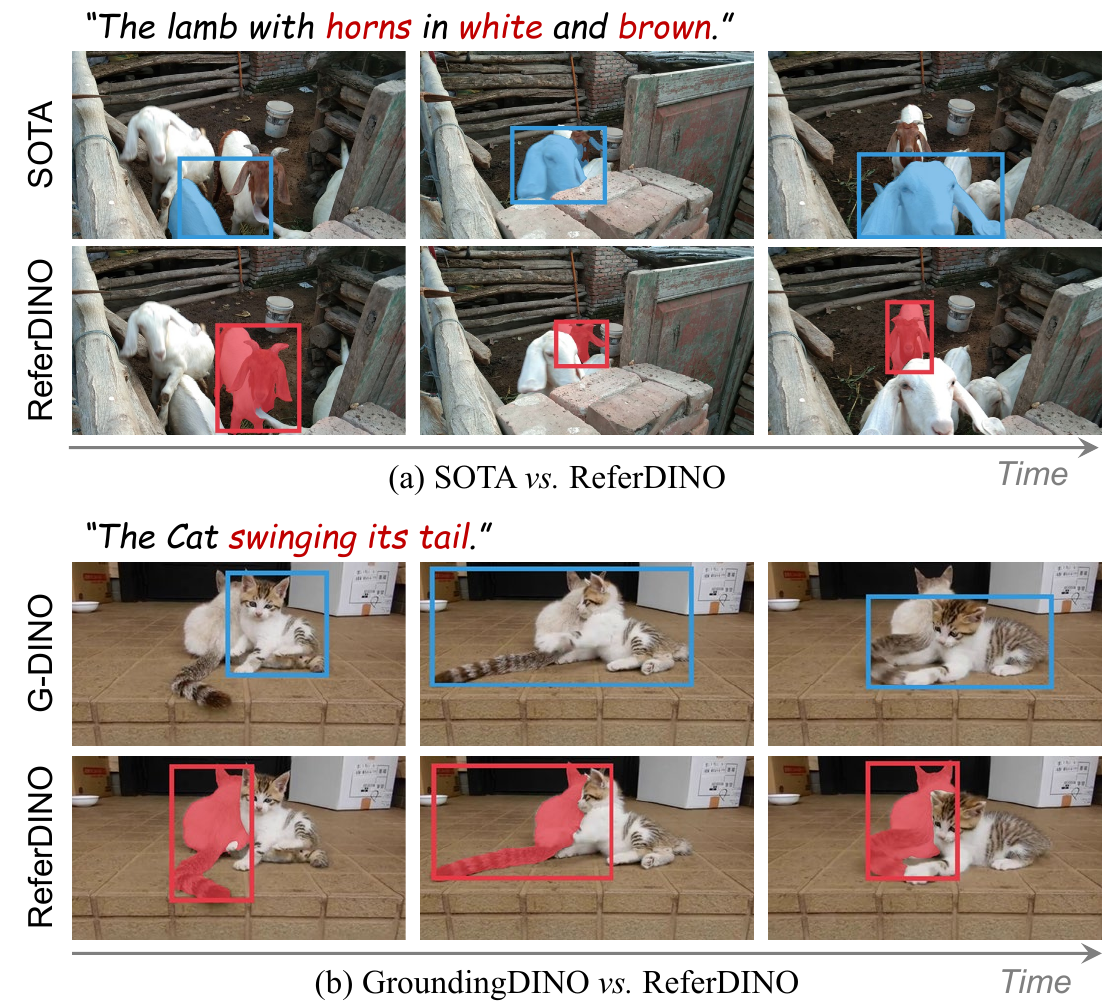}
    \vspace{-.5em}
    \caption{\textbf{(a)} The SOTA method~\cite{luo2023soc} in RVOS struggles to distinguish similar objects properly based on compositional attribute descriptions (e.g., shapes + colors), while our ReferDINO overcomes this limitation.
    \textbf{(b)} GroundingDINO~\cite{liu2024grounding} fails to identify the target cat based on the motion reference ``swinging its tail'', while our ReferDINO corrects the prediction through cross-modal spatiotemporal reasoning and generates precise object masks.
    }
    \vspace{-1em}
    \label{fig:compare}
 \end{figure}

Despite significant progress in recent years, existing RVOS models~\cite{botach2022end,wu2022language,luo2023soc,miao2023spectrum,yuan2024losh,he2024decoupling} still suffer from many common problems. 
For example, they often struggle to handle the queries involving complex appearance, relative locations and attributes, as shown in Figure~\ref{fig:compare} (a). These problems are mainly caused by the insufficient vision-language capabilities of current models, which, in turn, result from the limited scale and diversity of available RVOS data, as densely annotating video-text data is extremely expensive.

The recent progress in visual grounding field~\cite{liu2024grounding,li2022grounded,cheng2024yolo} offers unique potential to address this data limitation. 
Benefiting from large-scale pretraining on image-text data, the representative foundational models (e.g., GroundingDINO~\cite{liu2024grounding}) showcase profound capabilities in object-level vision-language understanding.
However, simply employing them to tackle RVOS task is impracticable, as shown in Figure~\ref{fig:compare} (b). 
There are mainly two challenges: 
(1) These models well-designed for region-level regression lack the ability of pixel-level dense prediction.
(2) Although they excel in understanding static attributes that can be observed in single frames (e.g., ``\textit{the white cat}''), they fail to identify the target object described by dynamic attributes (e.g., ``\textit{the cat swinging its tail}''). 

To address these challenges, we propose \textbf{ReferDINO}, a strong end-to-end RVOS approach that inherits static object perception from GroundingDINO~\cite{liu2024grounding}, and is further endowed with pixel-wise dense segmentation and spatio-temporal reasoning capabilities.
\underline{First}, we elaborate a \textit{grounding-guided deformable mask decoder} to produce high-quality object masks on each frame. 
Instead of simply adding a mask prediction branch in parallel with the original box prediction branch~\cite{li2023mask,wu2022language,luo2023soc,he2024decoupling}, our mask decoder cascades the two branches as a \textit{grounding-deformation-segmentation} pineline. 
It utilizes the pretrained box prediction as location prior, progressively refining the mask prediction through deformable attention mechanisms.
This process is differentiable, enabling mask learning to feed back into the box prediction branch for collaborative task learning.
\underline{Second}, we present an \textit{object-consistent temporal enhancer} that injects pretrained time-varying text features into inter-frame interaction for cross-modal temporal reasoning.
These two modules work together to bridge the gap between visual grounding and RVOS, effectively overcome the limitations of both foundational visual grounding models and existing RVOS methods, as shown in Figure~\ref{fig:compare}. 

Another limitation of these foundational models is their huge computational overhead, which is commonly unacceptable for video-scale training and inference.
To address this, we further propose a \textit{confidence-aware query pruning strategy} to reduce per-frame computations without compromising pretrained knowledge, by progressively identifying and pruning the low-confidence object queries.

Extensive experiments on five public RVOS benchmarks demonstrate that ReferDINO significantly outperforms state-of-the-art (SOTA) methods.
Moreover, compared to other baselines built upon GroundingDINO, our approach achieves remarkable performance improvements, with +\textbf{2.6 $\sim$ 3.6}\% \(\mathcal{J}\&\mathcal{F}\) on the Ref-Youtube-VOS dataset and 
and +\textbf{1.4 $\sim$ 4.6}\% \(\mathcal{J}\&\mathcal{F}\) on the MeViS.
In addition, our query pruning strategy reduces \textbf{40.6}\% FLOPs and \textbf{41.3}\% memory usage, achieving comparable performance with a \textbf{10$\times$} speedup (\textbf{4.9 $\rightarrow$ 51.0} FPS). These results demonstrate ReferDINO's potential for real-time video applications.

Overall, our contributions are summarized as follows:
\begin{itemize}
    \item We propose ReferDINO, a strong RVOS approach that inherits static object perception from pretrained foundational models and extends their capabilities to pixel-wise dense segmentation and spatiotemporal reasoning through a grounding-guided deformable mask decoder and an object-consistent temporal enhancer.
    \item We introduce a confidence-aware query pruning strategy, which significantly improves training and inference efficiency without compromising performance.
    \item ReferDINO establishes new SOTA performance on five benchmarks, and our proposed components provide significant improvements in performance and efficiency.
\end{itemize}

\section{Related Works}~\label{sec:rel_work}

\begin{figure*}[t]
    \centering
    \includegraphics[scale=0.38]{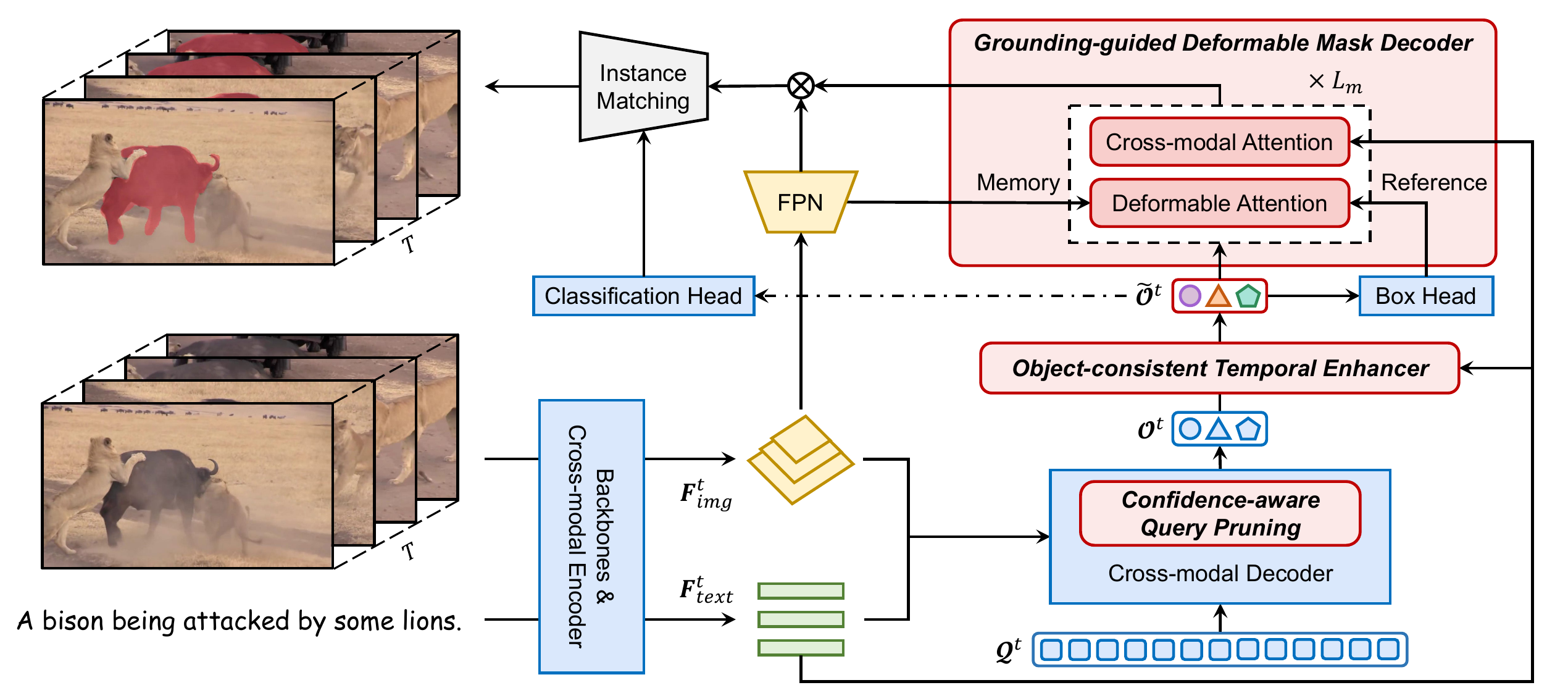}
    \vspace{-1em}
    \caption{Overall architecture of ReferDINO. Modules colored in \textcolor{mblue}{\textbf{blue}} are borrowed from GroundingDINO, while those in \textcolor{mred}{\textbf{red}} are newly introduced in this work. 
    First, a \textit{confidence-aware query pruning} strategy is employed to progressively prune low-confidence queries, deriving only a compact set of important object features $\mathcal{O}^t$. 
    After collecting the object features from all frames $\{{\mathcal{O}}^t\}_{t=1}^T$, the \textit{object-consistent temporal enhancer} performs cross-modal temporal reasoning by injecting time-varying text embeddings.
    Finally, our \textit{grounding-guided deformable mask decoder} takes the box predictions as location priors to refine the mask features through deformation attention and cross-modal attention mechanisms. Note that this process is end-to-end differentiable, enabling segmentation gradients to propagate backward into the box head for collaborative learning. Best viewed in color.}
    \vspace{-1em}
    \label{fig:referdino}
 \end{figure*}

\noindent
\textbf{Referring Video Object Segmentation.} RVOS~\cite{gavrilyuk2018actor,ding2023mevis,seo2020urvos} aims to segment objects throughout the video based on text descriptions.
MTTR~\cite{botach2022end} firstly introduces the DETR paradigm~\cite{carion2020end} into RVOS. Furthermore, ReferFormer~\cite{wu2022language} proposes to produce the queries from the text description.
On the top of this pipeline, follow-up works~\cite{han2023html,miao2023spectrum,luo2023soc,yuan2024losh} focus on modular improvements.
Despite notable progress on specific datasets, these models are limited by insufficient vision-language understanding, and often struggle in unseen objects or scenarios.
Recently, some works~\cite{grounded-sam-2,huang2024unleashing} attempt to utilize GroundingDINO to identify objects in single frames and then apply SAM2~\cite{ravi2024sam2} to generate dense masks.
However, such a manner of model ensemble is inefficient and non-differentiable, preventing further refinement of models' task-specific capabilities.
In contrast, our ReferDINO is an end-to-end adaptation approach that benefits from both open-world knowledge of GroundingDINO and specific knowledge from RVOS data.

\noindent
\textbf{Visual Grounding Foundational Model.}
Visual grounding~\cite{hong2019learning,hu2017modeling,liao2020real,sadhu2019zero,wang2018learning,wang2019neighbourhood,lin2025panoptic,wang2025pargo} aims to localize object regions in an image for a given text.
Recent works~\cite{li2022grounded,cheng2024yolo,yao2022detclip,liu2024grounding} focus on unifying detection and image-text datasets to increase the training vocabulary at scale. 
For example, GroundingDINO~\cite{liu2024grounding} incorporates grounded pretraining into DINO~\cite{zhang2022dino} to achieve impressive performance.
These models showcase strong object-level vision-language understanding, and have been explored in many downstream works~\cite{wasim2024videogrounding,pi2023detgpt,yang2024depth,zhang2024recognize,ren2024grounded}.  However, in most of these works, foundational models are primarily used as preprocessing tools to extract object regions, and their further potential in end-to-end task adaptation remains largely under explored.
Recently, Video-GroundingDINO~\cite{wasim2024videogrounding} extend GroundingDINO to produce temporal boundaries by simply inserting several temporal self-attention modules.
To the best of our knowledge, our work is the first end-to-end approach that adapts GroundingDINO to RVOS task. Besides, our designs upon foundational models can provide insights for other downstream fields.


\section{Background: GroundingDINO}
\label{sec:groundingdino}
Our approach builds on a visual grounding foundational model, GroundingDINO~\cite{liu2024grounding}, which we briefly revisit here. 
GroundingDINO is a DETR-based object detector, which introduces language into an object detector to achieve visual grounding. 
It mainly consists of an image backbone, a text backbone, a cross-modal encoder-decoder Transformer architecture, a box head, and a classification head. 

Given an image-text pair, GroundingDINO adopts the dual backbones to extract vanilla features, which are then fed into the cross-modal encoder to derive enhanced image features $\boldsymbol{F}_{\text{img}}$ and text features $\boldsymbol{F}_{\text{text}}$.
These enhanced cross-modal features are used to initialize a large set of query embeddings $\mathcal{Q}$, which are individually fed into a cross-modal decoder to generate object features $\mathcal{O}$. Note that the shapes of $\mathcal{O}$ and $\mathcal{Q}$ are equal.
Finally, each object feature is passed to the box head and the classification head to predict bounding boxes and scores.
Here, the scores are defined as the similarities between each object feature and the text tokens.
For RVOS, we define the binary classification probability as the maximum score over all tokens.

\section{ReferDINO}
\label{sec:referdino}
We illustrate our ReferDINO in Figure~\ref{fig:referdino}.
Apart from GroundingDINO, our model consists of three main components: grounding-guided deformable mask decoder (\S \ref{sec:head}), object-consistent temporal enhancer (\S \ref{sec:aggregator}) and confidence-aware query pruning (\S \ref{sec:selection}).
First, given a $T$-frame video and a language query, GroundingDINO is separately applied to each frame to obtain the object features.
To improve per-frame efficiency, we use the \textit{confidence-aware query pruning} strategy to progressively remove low-confidence object queries, achieving a very compact set of object features.
Then, collecting the object features of all frames, we utilize \textit{object-consistent temporal enhancer} to perform cross-modal temporal reasoning and ensure temporal consistency.
After that, we employ the \textit{grounding-guided deformable mask decoder}, which takes each object's box prediction as location condition, progressively refining its feature through the deformable cross-attention and cross-modal attention. 
Finally, the output feature corresponding to each object is used to produce a mask sequence $\{\mathbf{m}^t\}^T_{t=1}$ by dot-producting with the per-frame feature maps.

\subsection{Grounding-guided Deformable Mask Decoder}
\label{sec:head}
This decoder takes an object feature as input and generates corresponding pixel-wise predictions on each frame.
Instead of simply adding a mask head in parallel with the original box head, our mask decoder formulates the box and mask predictions as a \textit{grounding-deformation-segmentation} pineline. 
It utilizes the box prediction as location prior to iteratively enhance the mask predictions, and incorporates the text prompts for further refinement.
Moreover, this process is differentiable, enabling mask learning to feed back into the box head for collaborative task learning.

For simplicity, we use $\tilde{\boldsymbol{o}}$ to represent an arbitrary object feature. 
Before the mask decoder, we feed $\tilde{\boldsymbol{o}}$ into the box head to predict its bounding box $\boldsymbol{b} \in \mathbb{R}^{4}$, where $\boldsymbol{b}=\{b_x, b_y, b_w, b_h \}$ encodes the normalized box center coordinates, box height and width. 
Meanwhile, we employ a feature pyramid network (FPN) upon $\boldsymbol{F}_{\text{img}}$, producing a high-resolution feature map $\boldsymbol{F}_{\text{seg}}\in \mathbb{R}^{\frac{H}{4} \times \frac{W}{4} \times d}$, where $H$ and $W$ denote the height and width of the raw video frames. 

Then, we input these features and box predictions into the mask decoder for location-aware object refinement and mask generation.
Our mask decoder consists of $L_m$ blocks, with each block comprising two components: deformable cross-attention~\cite{zhu2020deformable} and cross-modal attention.
When applying attention on image feature maps (termed \textit{memory}), deformable cross-attention aggregates the features by adaptively sampling a small set of sampling points around a \textit{reference point}.
Unlike the vanilla deformable cross-attention~\cite{zhu2020deformable} that generates reference points through MLPs, we directly use the predicted box center $\{b_x, b_y\}$ as the reference point. Specifically, we treat $\tilde{\boldsymbol{o}}$ as the \textit{query}, $\boldsymbol{F}_{\text{seg}}$ as the \textit{memory}, and the normalized box center $\{b_x, b_y\}$ as the \textit{reference point}.
The sampling process is implemented by \textit{bilinear interpolation}~\cite{kirkland2010bilinear}, thus it is end-to-end differentiable.
In this way, the pretrained grounding knowledge is naturally incorporated to refine the mask prediction, and the segmentation gradients are also backward to optimize the object grounding.
A concern is that noisy features (e.g. background) can also be involved during adaptive sampling, which potentially compromises the mask quality. 
To mitigate this, we further apply cross-modal attention to integrate the text conditions, by taking $\tilde{\boldsymbol{o}}$ as \textit{query} and  $\boldsymbol{F}_{\text{text}}$ as \textit{key} \& \textit{value}.
These two attention mechanisms work together to ensure that mask prediction is tightly coupled with both text prompts and object locations.
Finally, for each object query, we obtain a refined mask embedding $\boldsymbol{o}_m \in \mathbb{R}^d$, which is then dot-producted with the high-resolution feature map $\boldsymbol{F}_{\text{seg}}$ to generate an instance mask $\boldsymbol{m}$. 

\noindent\textbf{Discussion.} The dynamic mask head~\cite{tian2020conditional} also integrates object location information into mask prediction by separately concatenating the feature map with each object's relative coordinates, and it has been adopted in many previous works~\cite{botach2022end,wu2022language,luo2023soc,wu2023onlinerefer,miao2023spectrum,yuan2024losh}.
However, the memory cost of this approach is dramatically high, due to the storage demand of per-object high-resolution feature maps.
This is particularly unacceptable for foundational models, which typically uses a large number of object queries.
In contrast, our mask decoder efficiently integrates the location information of different objects by location-guided sampling on a shared feature map. 
This design requires no extra memory burden, making our mask decoder particularly suitable for adaptation of foundational models.

\subsection{Object-consistent Temporal Enhancer}
\label{sec:aggregator}
\begin{figure}[t]
    \centering
    \includegraphics[height=4.5cm]{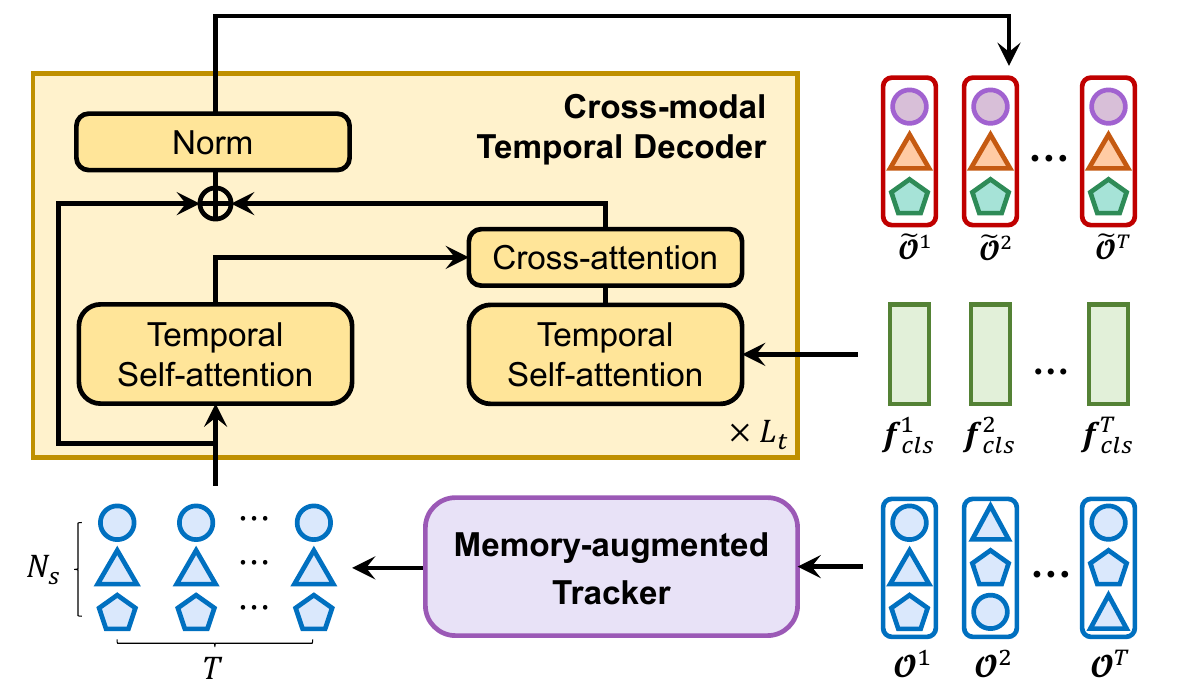}\vspace{-.5em}
    \caption{Illustration of our \textit{object-consistent temporal enhancer}, where $f^t_{cls}$ is the cross-modal sentence feature of $t$-th frame.}
    \vspace{-1em}
    \label{fig:temporal}
 \end{figure}

Although GroundingDINO can detect referring objects from single images, this is not reliable enough for RVOS.
First, it fails to understand the dynamic attributes in descriptions (e.g., action or temporal relationship).
Second, video frames often contain camera motion blur and constrained perspectives, greatly undermining its temporal consistency.
Therefore, we introduce an \textit{object-consistent temporal enhancer} to enable corss-modal temporal reasoning.
As shown in Figure~\ref{fig:temporal}, this module consists of a memory-augmented tracker and a cross-modal temporal decoder.
It receives two inputs: all the object embeddings $\{{\mathcal{O}}^t\}_{t=1}^T$, and time-varying sentence features $\{\boldsymbol{f}_{\text{cls}}^t\}_{t=1}^T$, where $\boldsymbol{f}_{\text{cls}}^t$ corresponds to the \texttt{[CLS]} token in $\boldsymbol{F}^t_{\text{text}}$.  

\noindent\textbf{Memory-augmented Tracker.}
Before temporal interaction, we align the objects across different frames with a tracker mechanism. 
Let $\mathcal{M}^t$ indicates the memory for $t$-th frame, and $\mathcal{M}^1={\mathcal{O}}^1$. 
Our tracker consists of two steps: \textit{object alignment} and \textit{memory updating}.
In the first step, we compute the cosine similarity between $\mathcal{M}^{t-1}$ and ${\mathcal{O}}^t$ as the assignment cost, and apply the Hungarian algorithm~\cite{kuhn1955hungarian} to align the objects with the memory:
\begin{equation}
    \hat{\mathcal{O}}^t = \operatorname{Hungarian}\left(\mathcal{M}^{t-1}, {\mathcal{O}}^t \right),
\end{equation}
where ${\hat{\mathcal{O}}}^t$ indicates the aligned object embeddings.
In the second step, these embeddings are used to update the memory in a momentum manner. 
Meanwhile, we incorporate the text relevance to prevent object-invisible frames from disturbing the long-term memory.
Formally, the memory is updated as follows: 
\begin{equation}
    \mathcal{M}^{t} = (1-\alpha \cdot \boldsymbol{c}^t) \cdot \mathcal{M}^{t-1} + \alpha\cdot \boldsymbol{c}^t \cdot \hat{\mathcal{O}}^t,
\end{equation}
where $\alpha$ is the momentum coefficient, and $\boldsymbol{c} \in \mathbb{R}^{N_s}$ is the cosine similarity between ${\hat{\mathcal{O}}}^t \in \mathbb{R}^{N_s \times d}$ and $\boldsymbol{f}_{\text{cls}}^t \in \mathbb{R}^{d}$.

\noindent\textbf{Cross-modal Temporal Decoder.} 
This module takes the time-varying text embeddings as frame proxies to perform inter-frame interaction and temporal enhancement.
Specifically, this module comprises $L_t$ blocks. 
In each block, given the aligned object embeddings $\{\hat{\mathcal{O}}^t\}_{t=1}^T$ and the sentence embeddings $\{\boldsymbol{f}^t_{\text{cls}}\}_{t=1}^T$, we employ self-attention along the temporal dimension to achieve inter-frame interaction.
Next, we extract dynamic information with a cross-attention module, which takes the sentence embeddings as \textit{query} and the object embeddings as \textit{key} and \textit{value}, deriving temporal-enhanced object features $\{\mathcal{O}_v^t\}_{t=1}^T$. 
These features containing effective temporal information are used to enhance the frame-wise object embeddings as follows: 
\begin{equation}
    \tilde{\mathcal{O}}^t= \operatorname{LayerNorm}\left(\hat{\mathcal{O}}^t + \mathcal{O}_v^t \right).
\end{equation}

\noindent\textbf{Discussion.} Our temporal enhancer improves existing temporal modules of RVOS models in two aspects.
1) Previous works either ignore object tracking~\cite{wang2021end,luo2023soc,miao2023spectrum,tang2023temporal} or only consider it in adjacent frames~\cite{ding2023mevis,wu2023onlinerefer,he2024decoupling}, while we employ a memory-augment tracker for stable long-term consistency.
2) In cross-modal temporal interaction, prior methods~\cite{wu2022language,luo2023soc,tang2023temporal,he2024decoupling} always use a static text embedding from the text encoder (e.g., BERT) for all objects across different frames, while our temporal enhancer uses time-varying text embeddings to better capture fine-grained temporal dynamics.
These two designs make our temporal enhancer fully exploit the pretrianed object knowledge and cross-modal features from foundational models for effective spatiotemporal reasoning and object consistency.

\subsection{Confidence-aware Query Pruning}
\begin{figure}[t]
    \centering
    \includegraphics[height=4.5cm]{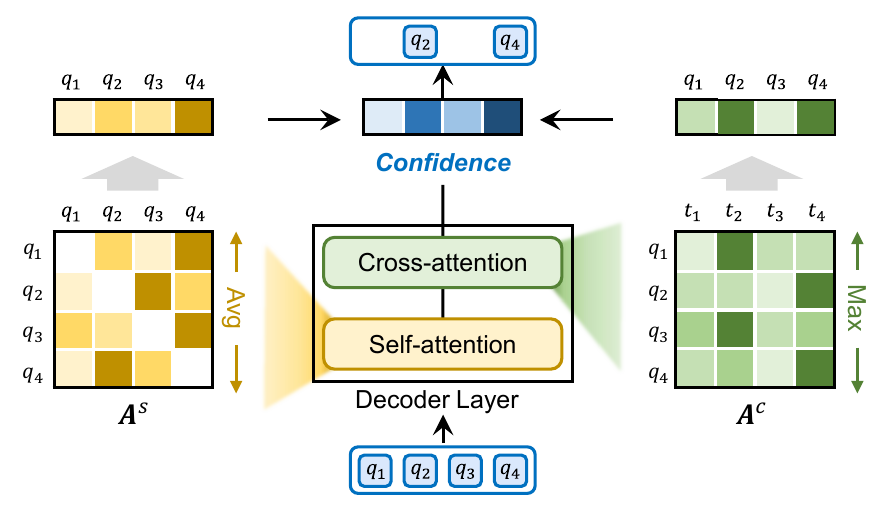}\vspace{-.5em}
    \caption{Illustration of our \textit{confidence-aware query pruning}, where $q_i$ and $t_i$ indicate $i$-th query and text token, respectively.}\vspace{-1em}
    \label{fig:selection}
 \end{figure}

Foundational visual grounding models typically utilize a large set of query embeddings to store extensive object information, e.g., GroundingDINO uses $N_q=900$ queries.
Iteratively processing such a large amount of queries significantly limits the efficiency, especially in video processing. 
However, directly reducing these queries can compromise the well-pretrained object knowledge. 
To solve this dilemma, we design a \textit{confidence-aware query pruning} strategy to progressly identify and reject low-confidence queries at each decoder layer, as shown in Figure~\ref{fig:selection}.

Specifically, the cross-modal decoder is stacked by $L$ layers, and each layer consists of a self-attention, a cross-attention with image features and a cross-attention with text features.
Let $\boldsymbol{Q}_l \in \mathbb{R}^{N_l\times d}$ denote the output query embeddings of the $l$-th decoder layer, where $N_0=N_q$. 
We reuse the attention weights in the decoder layer to compute a confidence score for each query as follows:
\begin{equation}
    \setlength{\abovedisplayskip}{.5em}
    s_j = \frac{1}{N_l-1} \sum_{\substack{i=1, i \neq j}}^{N_l} \boldsymbol{A}^s_{ij} + \max_k \boldsymbol{A}^c_{kj},  
    \setlength{\belowdisplayskip}{.5em}
\end{equation}
where $s_j$ is the confidence of $j$-th query, $\boldsymbol{A}^s \in \mathbb{R}^{N_l \times N_l}$ denotes the self-attention weight, $\boldsymbol{A}^c \in \mathbb{R}^{K \times N_l}$ denotes the transposed cross-attention weight with $K$ text tokens.
The former term represents the average attention received by $j$-th query from other queries. A query receiving high attention from others typically implies that it is irreplaceable. The latter maximum term measures the probability that $j$-th object query is mentioned in the text.
The combination of these two terms comprehensively reflects the query importance. 
Based on this score, we retain only $1/k$ high-confidence object queries at each layer, finally yielding a compact set of $N_s$ object embeddings, where $N_s \ll N_q$.
This strategy significantly reduces the computation costs while preserving the pretrianed object knowledge, enabling ReferDINO to achieve real-time efficiency.

\noindent\textbf{Time Complexity Analysis.} Let $N=N_q$, total time complexity of the original decoder is $O\left(L(N^2d+Nd^2)\right)$. If we retain only $1/k$ queries per layer, the total query number would decrease exponentially, and the total time complexity becomes $O\left((\frac{k^2}{k^2-1})N^2d + (\frac{k}{k-1})Nd^2\right)$, independent of the decoder depth $L$. In practice, this computational improvement is significant.
For GroundingDINO with $N=900$, $L=6$, $d=256$, even when $k=2$, the decoder computational cost can be reduced to 24.7\%.
More detailed derivation is presented in the supplementary~\ref{sup_derivation}.

\label{sec:selection}

\subsection{Training and Inference}
\label{sec:train_inference}
ReferDINO ultimately produces $N_s$ object prediction sequences $\boldsymbol{p} =\{\boldsymbol{p}_i \}_{i=1}^{N_s} $ for a video-text pair, and each sequence is represented by $\boldsymbol{p}_i = \{\boldsymbol{c}^t_i, \boldsymbol{b}^t_i, \boldsymbol{m}^t_i\}_{t=1}^{T}$, which denotes the binary classification probability, bounding box and mask of the $i$-th object query on $t$-th frame.

\renewcommand{\arraystretch}{1}
\begin{table*}[t]
    \setlength{\tabcolsep}{8pt}
    \centering
    \resizebox{\textwidth}{!}{
    \begin{tabular}{l | c |c c c | c c c| c c c| c c c}
        \hline
        \multirow{2}{*}{Method} & \multirow{2}{*}{Venue} &  \multicolumn{3}{c |}{Ref-YouTube-VOS} & \multicolumn{3}{c|}{Ref-DAVIS17} &  \multicolumn{3}{c |}{A2D-Sentences} & \multicolumn{3}{c}{JHMDB-Sentences} \\
         & & \( \mathcal{J} \)\&\( \mathcal{F} \) & \( \mathcal{J} \) & \( \mathcal{F} \)  &  \( \mathcal{J} \)\&\( \mathcal{F} \) & \( \mathcal{J} \) & \( \mathcal{F} \) & mAP & oIoU & mIoU  &  mAP & oIoU & mIoU \\
        \hline 
        \hline
        \multicolumn{14}{c}{\textit{Video-Swin-T / Swin-T}}\\
        \hline
        ReferFormer~\cite{wu2022language} & CVPR'22 & 59.4 & 58.0 & 60.9  & 59.6 & 56.5 & 62.7 & 52.8 & 77.6 & 69.6 & 42.2 & 71.9 & 71.0 \\
        HTML~\cite{han2023html} & ICCV'23 & 61.2 &59.5 &63.0&-&-&- & 53.4 & 77.6 &69.2 &42.7&-&-\\
        SgMg~\cite{miao2023spectrum}  & ICCV'23 & {62.0} & {60.4} & {63.5}  & {61.9} & {59.0} & {64.8} & {56.1} & {78.0} & {70.4} & {44.4} & {72.8} & {71.7}\\ 
        SOC~\cite{luo2023soc} & NeurIPS'23 &62.4& 61.1& 63.7 &63.5 &60.2 &66.7&54.8 &78.3& 70.6 &42.7 &72.7 &71.6\\
        LoSh~\cite{yuan2024losh} & CVPR'24 &63.7& 62.0& 65.4 &62.9 &60.1 &65.7&57.6& 79.3& 71.6 & - & - &-\\
        DsHmp~\cite{he2024decoupling} & CVPR'24 &63.6& 61.8& 65.4 &64.0 &60.8 &67.2&57.2& 79.0& 71.3 &44.9 &73.1 &72.1\\
        \hline
        \rowcolor{gray!20} G-DINO+SH & - & 63.9 & 62.0 & 65.8 & 62.0 & 58.3 & 65.9 & 57.8 & 79.0 & 71.0 & 44.6 & 72.7 & 71.6 \\
        \rowcolor{gray!20} G-DINO+DH & - & 64.2 & 62.4 & 66.1 & 65.1 & 61.3 & 69.1 & 57.3 & 78.2 & 69.9 & 45.1 & 73.2 & 71.8 \\
        \hline
        \rowcolor{red!10} \textbf{ReferDINO (ours)} & ICCV'25 &\textbf{67.5}& \textbf{65.5}& \textbf{69.6} &\textbf{66.7} &\textbf{62.9} &\textbf{70.7}&\textbf{58.9}& \textbf{80.2}& \textbf{72.3} &\textbf{45.6} &\textbf{74.2} &\textbf{73.1}\\
        \hline
        \hline
        \multicolumn{14}{c}{\textit{Video-Swin-B / Swin-B}}\\
        \hline
        ReferFormer~\cite{wu2022language} & CVPR'22 & 62.9 & 61.3 & 64.6 & 61.1 & 58.1 & 64.1 & 55.0 & 78.6 & 70.3 & 43.7 & 73.0 & 71.8\\
        HTML~\cite{han2023html} & ICCV'23 & 63.4 &61.5 &65.2 &62.1 &59.2 &65.1& 56.7 & 79.5 &71.2 &44.2&-&-\\
        SgMg~\cite{miao2023spectrum}  & ICCV'23 &  {65.7} & {63.9} & {67.4}  & {63.3} & {60.6} & {66.0} &  {58.5} & {79.9} & {72.0} & {45.0} & {73.7} & {72.5}\\ 
        SOC~\cite{luo2023soc} & NeurIPS'23 &66.0 &64.1& 67.9& 64.2 &61.0 &67.4&57.3  &80.7 & 72.5 & 44.6 &73.6 &72.3\\
        LoSh~\cite{yuan2024losh} & CVPR'24 &67.2& 65.4& 69.0 &64.3 &61.8 &66.8&59.9& 81.2& 73.1 & - & - &-\\
        DsHmp~\cite{he2024decoupling} & CVPR'24 &67.1& 65.0& 69.1 &64.9 &61.7 &68.1&59.8& 81.1& 72.9 &45.8 &73.9 &73.0\\
        \hline 
        \rowcolor{gray!20} G-DINO+SH & - & 66.2 & 64.2 & 68.3 & 65.2 & 61.6 & 68.8 & 58.4 & 80.9 & 72.6 & 45.5 & 74.2 & 72.7 \\
        \rowcolor{gray!20} G-DINO+DH & - & 66.7 & 64.5 & 69.0 & 67.6 & 63.6 & 71.3 & 58.7 & 80.2 & 71.8 & 45.3 & 74.1 & 73.2 \\
        \hline
        \rowcolor{red!10} \textbf{ReferDINO (ours)} & ICCV'25 &\textbf{69.3}& \textbf{67.0}& \textbf{71.5} &\textbf{68.9} &\textbf{65.1} &\textbf{72.9}&\textbf{61.1}& \textbf{82.1}& \textbf{73.6} &\textbf{46.6} &\textbf{74.2} &\textbf{73.2}\\
        \hline 
    \end{tabular}}
\vspace{-1em}
\caption{Comparison on Ref-YouTube-VOS, Ref-DAVIS17, A2D-Sentences and JHMDB-Sentences. 
}
\vspace{-1.5em} 
\label{tab:ytvos_davis}
\end{table*}

\noindent\textbf{Training.} Suppose the ground truth object sequence to be $\boldsymbol{y} = \{\boldsymbol{c}^t, \boldsymbol{b}^t, \boldsymbol{m}^t\}_{t=1}^{T}$.
Following the previous works~\cite{carion2020end,botach2022end,luo2023soc}, we apply the Hungarian algorithm for instance matching. 
Specifically, we select the instance sequence with the lowest matching cost as the positive and assign the remaining sequences as negative. The matching cost is defined as:
\begin{equation}
    \begin{aligned}
    \mathcal{L}_\text{total}\left(\boldsymbol{y}, \boldsymbol{p}_i\right) &= \lambda_{\text{cls}} \mathcal{L}_{\text{cls}}\left(\boldsymbol{y}, \boldsymbol{p}_i \right) + \lambda_{\text{box}} \mathcal{L}_{\text{box}}\left(\boldsymbol{y}, \boldsymbol{p}_i \right) \\ &+ \lambda_{\text{mask}} \mathcal{L}_{\text{mask}}\left(\boldsymbol{y}, \boldsymbol{p}_i \right).
    \end{aligned}
\end{equation}
This matching cost is computed on individual frames and normalized by the frame number.
Here, $\mathcal{L}_{\text{cls}}$ is the focal loss that supervises the binary classification prediction. $\mathcal{L}_{\text{box}}$ sums up the L1 loss and GIoU loss. $\mathcal{L}_{\text{mask}}$ is the combination of DICE loss, binary mask focal loss and projection loss~\cite{tian2020conditional}.
$\lambda_{\text{cls}}$, $\lambda_{\text{box}}$ and $\lambda_{\text{mask}}$ are scalar weights of individual losses.
The model is optimized end-to-end by minimizing the total loss $\mathcal{L}_\text{total}$ for positive sequences and only the classification loss $\mathcal{L}_{\text{cls}}$ for negative sequences.

\noindent\textbf{Inference.} At inference, we select the best sequence with the highest average classification score as follows:
\begin{equation}
    \label{eq:infer}
    \setlength{\abovedisplayskip}{.5em}
    \sigma=\underset{i \in [1, N_s]}{\arg \max } \frac{1}{T} \sum\nolimits_{t=1}^T \boldsymbol{c}^t_i.
    \setlength{\belowdisplayskip}{.5em}
\end{equation}
Finally, the output mask sequence is formed as $\{\boldsymbol{m}_\sigma^t \}_{t=1}^T$.

\section{Experiments}
\subsection{Experimental Setup}
\textbf{Datasets and Metrics.} 
We evaluate ReferDINO on five public RVOS benchmarks: 
Ref-YouTube-VOS~\cite{seo2020urvos} covers 3,978 videos with 15K descriptions.
Ref-DAVIS17~\cite{khoreva2019video} contains 90 videos and 1.5K descriptions.
A2D-Sentences~\cite{gavrilyuk2018actor} contains 3.7K videos and 6.6K descriptions.
JHMDB-Sentences~\cite{gavrilyuk2018actor} contains 928 videos and 928 descriptions.
MeViS~\cite{ding2023mevis} contains 2K videos and 28K descriptions.  
On Ref-YouTube-VOS, Ref-DAVIS17 and MeViS, we use region similarity $\mathcal{J}$,
contour accuracy $\mathcal{F}$, and their average $\mathcal{J\&F}$ as metrics. 
For A2D-Sentences and JHMDB-Sentences, we employ mAP, overall IoU (oIoU), and mean IoU (mIoU) metrics.

\noindent
\textbf{Protocols.}
We follow the protocols established in previous works, and all evaluations are conducted using the official code or online platforms. Specifically, the model trained on the training set of Ref-YouTube-VOS is directly evaluated on both Ref-YouTube-VOS and Ref-DAVIS17. Similarly, the model trained on A2D-Sentences is directly evaluated on A2D-Sentences and JHMDB-Sentences. 
The model is first pretrained on image datasets RefCOCO/+/g~\cite{kazemzadeh2014referitgame, mao2016generation}, and then trained on the RVOS datasets, except for MeViS where the model is trained directly, following \cite{ding2023mevis,he2024decoupling}.

\noindent
\textbf{Implementation Details.}
Our model is built upon the pretrained GroundingDINO, which uses Swin Transformer as the image backbone and BERT as the text backbone. The official source releases two GroundingDINO checkpoints: Swin-T and Swin-B, both of which are covered in our experiments.
We freeze the backbones and finetune the cross-modal Transformer with LoRA techniques~\cite{hu2021lora}, where the rank is set to $32$.
We set $\alpha=0.1$, $L_t=3$, and $L_m=3$. 
In the mask decoder, we set the number of sampling points to 16.
In the query pruning, we set the drop rate as $50\%$. 
For MeViS that involves multiple target objects, we follow the practice in \cite{ding2023mevis,he2024decoupling} to select multiple object trajectories with classification scores higher than a threshold $\sigma=0.3$. 
For the other datasets, we select the best object trajectory.
Extra training details can be found in the supplementary \ref{sup_details}.

\renewcommand{\arraystretch}{1}
\begin{table}[t]
    \setlength{\tabcolsep}{11pt}
    \centering
    \resizebox{.48\textwidth}{!}
    {\begin{tabular}{l|c|ccc}
        \hline
        Method & Venue & $\mathcal{J}$\&$\mathcal{F}$ & $\mathcal{J}$ & $\mathcal{F}$ \\
        \hline
        \hline
        \multicolumn{5}{c}{\textit{Video-Swin-T / Swin-T}}\\
        \hline
        MTTR~\cite{botach2022end} & CVPR'22 & 30.0 & 28.8 & 31.2 \\
        ReferFormer~\cite{wu2022language} & CVPR'22 & 31.0 & 29.8 & 32.2 \\
        LMPM~\cite{ding2023mevis} & ICCV'23 & 37.2 & 34.2 & 40.2 \\
        DsHmp~\cite{he2024decoupling} & CVPR'24 & 46.4 & 43.0 & 49.8 \\
        \hline
        \rowcolor{gray!20} G-DINO-SH & - & 43.4 & 39.8 & 47.1 \\
        \rowcolor{gray!20} G-DINO-DH & - & 45.4 & 41.9 & 48.9 \\
        \hline
        \rowcolor{red!10} \textbf{ReferDINO (ours)} & ICCV'25 & \textbf{48.0} & \textbf{43.6} & \textbf{52.3} \\
        \hline
        \hline
        \multicolumn{5}{c}{\textit{Video-Swin-B / Swin-B}}\\
        \hline
        \rowcolor{gray!20} G-DINO-SH & - & 45.0 & 41.3 & 48.9 \\
        \rowcolor{gray!20} G-DINO-DH & - & 47.9 & 44.3 & 51.5 \\
        \hline
        \rowcolor{red!10} \textbf{ReferDINO (ours)} & ICCV'25 & \textbf{49.3} & \textbf{44.7} & \textbf{53.9} \\
        \hline
    \end{tabular}}
    \vspace{-1em}
    \caption{Comparison on MeViS.} \vspace{-2em}
    \label{tab:MeViS}
\end{table}

\subsection{Main Results}

\textbf{Comparisons with SOTA Methods.} As shown in Tables~\ref{tab:ytvos_davis} and \ref{tab:MeViS}, our ReferDINO significantly and consistently outperforms SOTA methods across all five RVOS datasets. Specifically, with the Swin-T backbone, our ReferDINO achieves 48.0\% \(\mathcal{J}\&\mathcal{F}\) on MeViS, outperforming existing SOTA by 1.6\%. On Ref-YouTube-VOS, our ReferDINO achieves 66.4\% \(\mathcal{J}\&\mathcal{F}\), improving SOTA performance by 2.8\%. 
When a larger backbone Swin-B is applied, ReferDINO further improves \(\mathcal{J}\&\mathcal{F}\) to 69.3\% on Ref-YouTube-VOS, surpassing SOTA by more than 2.2\%. Consistent performance improvements are observed across the other datasets, which demonstrates the superiority of ReferDINO.

\noindent\textbf{Comparisons with Baselines.} To further evaluate ReferDINO's effectiveness, we develop two enhanced variants of GroundingDINO as baselines: G-DINO+SH (static mask head) and G-DINO+DH (dynamic mask head~\cite{tian2020conditional}). 
The former directly dot products the query embedding with the feature map, while the latter uses object-wise feature maps as discussed in Section~\ref{sec:head}.
These two mask heads are generic and commonly used in previous RVOS models~\cite{botach2022end,wu2022language,luo2023soc,miao2023spectrum,yuan2024losh,he2024decoupling}.
Moreover, both baselines are augmented with temporal self-attention modules for cross-frame interaction, as in VideoGroundingDINO~\cite{wasim2024videogrounding}.
As shown in Tables~\ref{tab:ytvos_davis} and ~\ref{tab:MeViS}, although these baselines achieve competitive results on specific datasets (e.g., on Ref-DAVIS17, G-DINO-SH surpasses SOTAs by +0.3\% \(\mathcal{J}\&\mathcal{F}\) and G-DINO-DH attains +2.7\%),
they fail to exhibit significant or consistent improvements over existing models across datasets.
These results demonstrate that trivial adaptation cannot fully unleash the pretrained capabilities of foundational models.
In contrast, our ReferDINO demonstrates robust superiority across different backbones and datasets.
The advantages become particularly significant on large benchmarks (Ref-Youtube-VOS and MeViS), where our ReferDINO with Swin-T can even outperform the baselines with Swin-B.
More comparison results of GroundingDINO baselines are presented in the supplementary~\ref{sup_comparison}.
These results demonstrate the effectiveness of our approach in adapting foundational models to RVOS task.

\subsection{Ablation Studies}
In this section, we mainly analyze the effectiveness of the three proposed components.
   
\noindent
\textbf{Mask Decoder.} 
Our \textit{grounding-guided deformable mask decoder} utilizes deformable cross-attention (DCA) for grounding-guided adaptive sampling and cross-modal attention (CMA) to further mitigate the impacts of sampling noise. To validate the effectiveness of these two modules, we conduct ablation studies.
As shown in Table~\ref{tab:seg_dec}, removing CA and DCA results in performance drops of 0.4\% and 2.7\% in \(\mathcal{J}\&\mathcal{F}\), respectively. 
These results demonstrate the complementary advantages of the two attention mechanisms, where DCA plays a dominant role in improving mask quality and CA provides further refinement.

 
\renewcommand{\arraystretch}{1}
\begin{table}[t]
    \setlength{\tabcolsep}{25pt}
    \centering
    \resizebox{.48\textwidth}{!}
    {\begin{tabular}{l|ccc}
        \hline
        Method & $\mathcal{J}$\&$\mathcal{F}$ & $\mathcal{J}$ & $\mathcal{F}$ \\
        \hline
        \rowcolor{red!10} \textbf{ReferDINO} & \textbf{48.0} & \textbf{43.6} & \textbf{52.3} \\
        \hline
        w/o CMA & 47.6 & 43.1 & 52.0 \\
        w/o DCA & 45.3 & 41.8 & 48.7  \\
        \hline
    \end{tabular}}
    \vspace{-1em}
    \caption{Ablation study of the mask decoder on MeViS.}
    \vspace{-.5em}
    \label{tab:seg_dec}
\end{table}

\noindent
\textbf{Temporal Enhancer.}
Our \textit{object-consistent temporal enhancer} consists of a memory-augmented tracker and a cross-modal temporal decoder. We analyze their individual contributions in Table~\ref{tab:temp}. First, while the object queries in GroundingDINO is naturally aligned across frames to some extent, adding a tracker can still improve the temporal consistency, leading to a 0.4\% improvement in \(\mathcal{J}\&\mathcal{F}\). 
Second, in the temporal decoder is essential for understanding temporal dynamics and object motion, providing 2.2\% improvements in \(\mathcal{J}\&\mathcal{F}\).
We additionally present visualizations to show the effects of memory-augmented tracker in the supplementary \ref{sup_ablation}.
These results demonstrates the effectiveness of our object-consistent temporal enhancer.

\begin{figure*}[htp]
    \centering
    \includegraphics[scale=0.48]{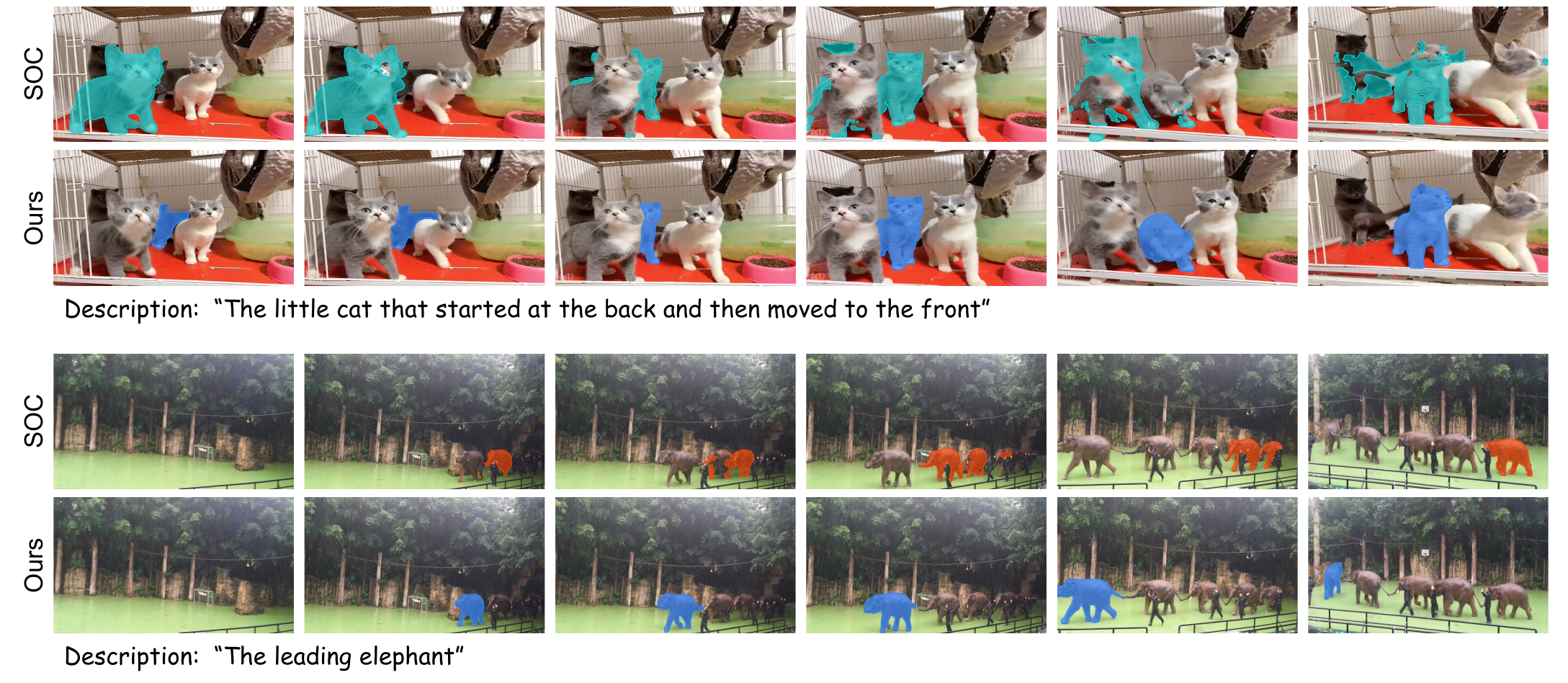}
    \vspace{-2em}
    \caption{Qualitative results of our ReferDINO with the SOTA method~\cite{luo2023soc}. ReferDINO performs much better in understanding complex object description involving motion and spatial relationships.}
    \vspace{-.5em}
    \label{fig:example}
 \end{figure*}

\begin{figure*}[t]
    \centering
    \includegraphics[scale=0.48]{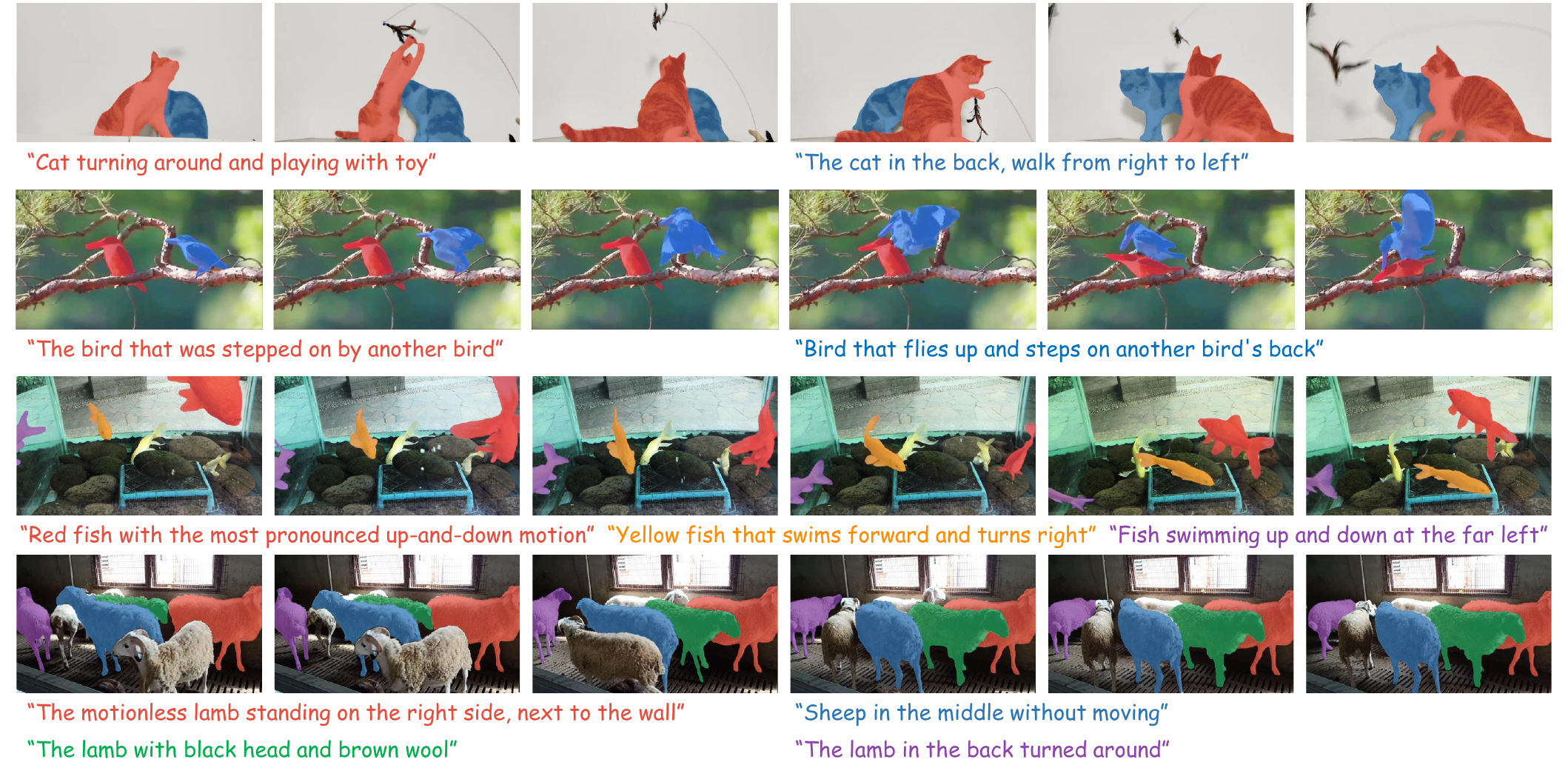}
    \vspace{-1em}
    \caption{Visualization of our ReferDINO for multiple text references.}
    \vspace{-1.5em}
    \label{fig:example_ours}
 \end{figure*}

\noindent
\textbf{Query Pruning.} We explore the impact of purning different numbers of queries. As training with more than 50\% of queries on MeViS is highly time-consuming, we conduct this validation on Ref-YouTube-VOS, and present the training costs, inference speed and performance in Table~\ref{tab:query}.
It shows that our query pruning strategy can effectively maintain the performance while significantly reducing the computational costs.
For instance, reducing 50\% queries with our strategy can dramatically reduce 40.6\% FLOPs and 41.2\% memory usage, making video-scale training on smaller GPUs possible. 
Meanwhile, our strategy achieves 51.0 FPS, increasing the inference speed by 10 $\times$ while maintaining consistent performance.
In contrast, randomly pruning the same number of queries results in a substantial 29.3\% decrease in \(\mathcal{J}\&\mathcal{F}\).
With our strategy, ReferDINO can be efficiently trained on large RVOS datasets and achieve real-time video processing.

\renewcommand{\arraystretch}{1}
\begin{table}[t]
    \setlength{\tabcolsep}{25pt}
    \centering
    \resizebox{.48\textwidth}{!}
    {\begin{tabular}{l|ccc}
        \hline
        Method & $\mathcal{J}$\&$\mathcal{F}$ & $\mathcal{J}$ & $\mathcal{F}$ \\
        \hline
        \rowcolor{red!10} \textbf{ReferDINO} & \textbf{48.0} & \textbf{43.6} & \textbf{52.3} \\
        \hline
        w/o Tracker & 47.6 & 43.2 & 52.1 \\
        w/o Decoder & 45.8 & 42.7 & 48.8 \\
        \hline
    \end{tabular}}
    \vspace{-1em}
    \caption{Ablation study of the temporal enhancer on MeViS.}
    \vspace{-.5em}
    \label{tab:temp}
\end{table}

\renewcommand{\arraystretch}{1}
\begin{table}[t]
    \setlength{\tabcolsep}{8pt}
    \centering
    \resizebox{.48\textwidth}{!}
    {\begin{tabular}{c|ccc|ccc}
        \hline
        Drop Rate & $\mathcal{J}$\&$\mathcal{F}$ & $\mathcal{J}$ & $\mathcal{F}$ & FLOPs & Memory & FPS \\
        \hline
        0 & 67.6 & 65.6 & 69.7 & 840.3G & 24.0G & 4.9 \\
        \hline
        15\% & 67.7 & 65.7 & 69.8 & 672.7G & 18.9G & 26.5 \\
        30\% & 67.6 & 65.6 & 69.6 & 572.4G & 15.9G & 47.2 \\
        \rowcolor{red!10} 50\% & 67.5 & 65.5 & 69.6 & 499.3G & 14.1G & 51.0 \\
        \hline
        \textit{Random} 50\% & 38.3 & 37.1 & 39.4 & 499.3G & 14.1G & 51.0 \\
        \hline
    \end{tabular}}
    \vspace{-1em}
    \caption{Ablation study of query pruning on Ref-YouTube-VOS.
    FLOPs and Memory are measured with batch size of $1$ on 6-frame 360P videos, and Pytorch checkpointing disabled.
    FPS is measured at 360P, following the official evaluation protocol~\cite{seo2020urvos}.
    }
    \vspace{-1.5em}    
    \label{tab:query}
\end{table}

\subsection{Qualitative Analysis}
In Figure~\ref{fig:example}, we provide the qualitative comparisons with the SOTA model~\cite{luo2023soc}. In Figure~\ref{fig:example_ours}, we present several visualizations of ReferDINO processing various text references within a video. 
More qualitative results are presented in the supplementary \ref{sup_visual}.
These scenarios involve multiple similar objects and complex object descriptions, including rich composite attributions, spatial relationships, and temporal motion. The results clearly demonstrate the effectiveness of ReferDINO in video object segmentation, spatio-temporal reasoning, and vision-language understanding.

\section{Conclusion}
In this work, we propose ReferDINO for RVOS task to address the challenges in vision-language understanding, dense perception and spatiotemporal reasoning.
ReferDINO inherits region-level vision-language alignment from GroundingDINO, and is further endowed with pixel-level object segmentation and cross-modal spatiotemporal reasoning capabilities, through a grounding-guided deformable mask decoder and an object-consistent temporal enhancer. 
Moreover, we propose a confidence-aware query pruning strategy to accelerate the decoding process.
Extensive experiments across five public benchmarks demonstrate that ReferDINO significantly outperforms existing RVOS methods.
Moreover, the proposed query pruning strategy significantly improves both training and inference efficiency without compromising performance.

{\footnotesize\vspace{.5em}\noindent\textbf{Acknowledgements}.
This work was supported partially by the NSFC (U22A2095, 6247072922), Guangdong Natural Science Funds Project (2023B1515040025), Guangdong NSF for Distinguished Young Scholar (2022B15-15020009), Guangdong Provincial Key Laboratory of Information Security Technology (2023B1212060026),  open research fund of Key Laboratory of Machine Intelligence and System Control, Ministry of Education (No. MISC-202407). We are grateful to the National Supercomputing Center in Guangzhou for providing computing resources.
}

{
    \small
    \bibliographystyle{ieeenat_fullname}
    \bibliography{main}
}

\clearpage
\appendix

\section*{Supplementary Material}

\section{Detailed Derivation.}\label{sup_derivation}
Here, we present the detailed derivation of time complexity for the decoder to demonstrate the effectiveness of our query pruning strategy.
GroundingDINO iteratively processes a large number of query embeddings, resulting in significant computational bottlenecks.
As GrounddingDINO is based on Deformable DETR~\cite{zhu2020deformable}, its decoder's computations are mainly occupied by the multi-head self-attention (MHSA) operations and feed-forward layers. MHSA is defined as follows:
\begin{equation}
    \operatorname{MHSA}(\mathbf{X}) = \sum_{i=1}^H \mathbf{W}^o_i \left[\operatorname{softmax}\left(\mathbf{Q}_i \mathbf{K}_i^T \right) \mathbf{V}_i\right], 
\end{equation}
\begin{equation}
    \text{and} \quad \mathbf{Q_i} = \mathbf{X}\mathbf{W}_i^q,\quad \mathbf{K}_i = \mathbf{X}\mathbf{W}_i^k,\quad \mathbf{V}_i = \mathbf{X}\mathbf{W}_i^v,
\end{equation}
where $\mathbf{X} \in \mathbb{R}^{N \times d}$ indicates the query embeddings, $H$ is the head number, and $\mathbf{W}_i^q$, $\mathbf{W}_i^k$, $\mathbf{W}_i^v$, $\mathbf{W}_i^o \in \mathbb{R}^{d\times d}$ are the linear weights of $i$-th head. The time complexity of each layer is $O(N^2d+Nd^2)$, hence the total time complexity for a $L$-layer decoder is $O\left(L(Nd^2+N^2d)\right)$. Then, if we employ the \textit{confidence-aware query pruning strategy} to identify and retain only $\frac{1}{k}$ queries at each layer, the total time complexity becomes: 
\begin{align}
    &O\left(\left[1+\frac{1}{k^2}+\frac{1}{k^4}+\cdots+\frac{1}{k^{2(L-1)}}\right]  N^2 d + \left[1+\right.\right. \nonumber \\
    &\quad\quad\quad\left.\left.\frac{1}{k}+\frac{1}{k^2}+\cdots+\frac{1}{k^{L-1}}\right]  Nd^2 \right) \\
    = &O\left(\frac{k^2}{k^2-1}  ({\color{mred}1-\frac{1}{k^{2L}}})  N^2d + \frac{k}{k-1}  ({\color{mred}1-\frac{1}{k^{L}}})  Nd^2\right) \label{eq:complexty} \\
    = &O\left(\frac{k^2}{k^2-1}  N^2d + \frac{k}{k-1}  Nd^2\right),
\end{align}
which is independent of the decoder depth $L$. Note that in equation~\ref{eq:complexty}, the items ${\color{mred}1-\frac{1}{k^{L}}}$ and ${\color{mred}1-\frac{1}{k^{2L}}}$ can be omitted because $\frac{1}{k^{2L}}$ and $\frac{1}{k^{L}}$ are typically much smaller than 1 when $k>1$ and $L$ is a relatively large positive integer.

\section{Additional Implementation Details.}
\textbf{FPN.} We follow the prior FPN structure~\cite{botach2022end,luo2023soc}, which consists of several 2D convolution, GroupNorm and ReLU Layers. Nearest neighbor interpolation is used for upsampling steps. 

\noindent\textbf{Implementation Details.} \label{sup_details}
The coefficients for losses are set as $\lambda_{cls}=4$, $\lambda_{L1}=5$, $\lambda_{giou}=2$, $\lambda_{dice}=5$, $\lambda_{focal}=5$, $\lambda_{proj}=5$. We apply image augmentation techniques such as flipping, rotation, cropping, and scaling during training, following  previous works~\cite{wu2022language, miao2023spectrum}. 
The model is pretrained on Ref-COCO/g/+ for 20 epochs with the batch size of 8 for Swin-T and 10 for Swin-B. Then we individually train the models on Ref-YouTube-VOS for 2 epochs and A2D-Sentence for 6 epochs.
For MeViS, we train the models for 15 epochs.
The number of sampling frames is 6.
We use AdamW as the optimizer with the weight decay of 1e-4 and initial learning rate of 5e-5, which is linearly rescaled with the batch size. 
All experiments are conducted on 8 NVIDIA A800 GPUs. 

\section{Comparisons with Vanilla Baselines.}\label{sup_comparison}
As mentioned in Section~\ref{sec:rel_work}, some previous works~\cite{grounded-sam-2,huang2024unleashing} also attempted to employ GroundingDINO for RVOS task. 
Grounded-SAM2~\cite{grounded-sam-2} extracts object regions with GroundingDINO and produces masks with SAM2~\cite{ravi2024sam2}.
Based on this, AL-Ref-SAM2~\cite{huang2024unleashing} further incorporates GPT4~\cite{achiam2023gpt} to select key frames and boxes.
However, such a manner of model ensemble is not end-to-end differentiable, preventing further refinement of RVOS-specific capability.
As shown in Table~\ref{sup_tab:MeViS}, our ReferDINO outperforms these ensemble methods by large margins on MeViS dataset, the largest RVOS benchmark.
These results demonstrate the advantages of our end-to-end adaptation approach.

\renewcommand{\arraystretch}{1}
\begin{table}[h]
    \setlength{\tabcolsep}{10pt}
    \centering
    \resizebox{.48\textwidth}{!}
    {\begin{tabular}{l|cccc}
        \hline
        Method & $\mathcal{J}$\&$\mathcal{F}$ & $\mathcal{J}$ & $\mathcal{F}$ & FPS \\
        \hline
        \hline
        \multicolumn{5}{c}{\textit{Video-Swin-T / Swin-T}}\\
        \hline
        Grounded-SAM2~\cite{grounded-sam-2} & 37.4 & 31.0 & 43.7 & 6.1 \\
        \hline
        \rowcolor{red!10} \textbf{ReferDINO (ours)} &  \textbf{48.0} & \textbf{43.6} & \textbf{52.3} & \textbf{28.0} \\
        \hline
        \hline
        \multicolumn{5}{c}{\textit{Video-Swin-B / Swin-B}}\\
        \hline
        Grounded-SAM2~\cite{grounded-sam-2} & 40.5 & 34.5 & 46.4 & 6.1  \\
        AL-Ref-SAM 2~\cite{huang2024unleashing} & 42.8 & 39.5 & 46.2 & $<$ 6.1  \\
        \hline
        \rowcolor{red!10} \textbf{ReferDINO (ours)} & \textbf{49.3} & \textbf{44.7} & \textbf{53.9} & \textbf{26.6} \\
        \hline
    \end{tabular}}
    \caption{Performance comparison on MeViS.} \vspace{-1em}
    \label{sup_tab:MeViS}
\end{table}

\section{Additional Latency Comparisons.}
In Table~\ref{sup_tab:latency}, we compare with SOTA dynamic-head methods and our baseline on Ref-Youtube-VOS.
The results show that our mask decoder reduces memory usage by 11.2G, enabling superior performance over SOTAs at comparable or faster speed.

\renewcommand{\arraystretch}{1}
\begin{table}[h]
    \setlength{\tabcolsep}{10pt}
    \centering
    \resizebox{.48\textwidth}{!}{
    \begin{tabular}{l|ccc}
        \hline 
        Model & $\cal{J}\&\cal{F}\uparrow$ & Memory$\downarrow$ & FPS$\uparrow$\\
        \hline
        SgMg~\cite{miao2023spectrum} & 62.0 & 16.5G & 50.3\\
        MUTR~\cite{yan2024referred} & 64.0 & 34.0G & 41.4\\\hline
        G-DINO+DH & 64.2 & 25.3G & 50.2\\\hline
        \textbf{ReferDINO (Ours)} & \textbf{67.5} & \textbf{14.1G} & \textbf{51.0}\\
      \hline
    \end{tabular}}
    \caption{Comparison of training memory and inference speed.}
    \label{sup_tab:latency}
\end{table}

\section{Additional Ablation Studies.}\label{sup_ablation}
\begin{figure}[h]
    \centering
    \includegraphics[scale=0.46]{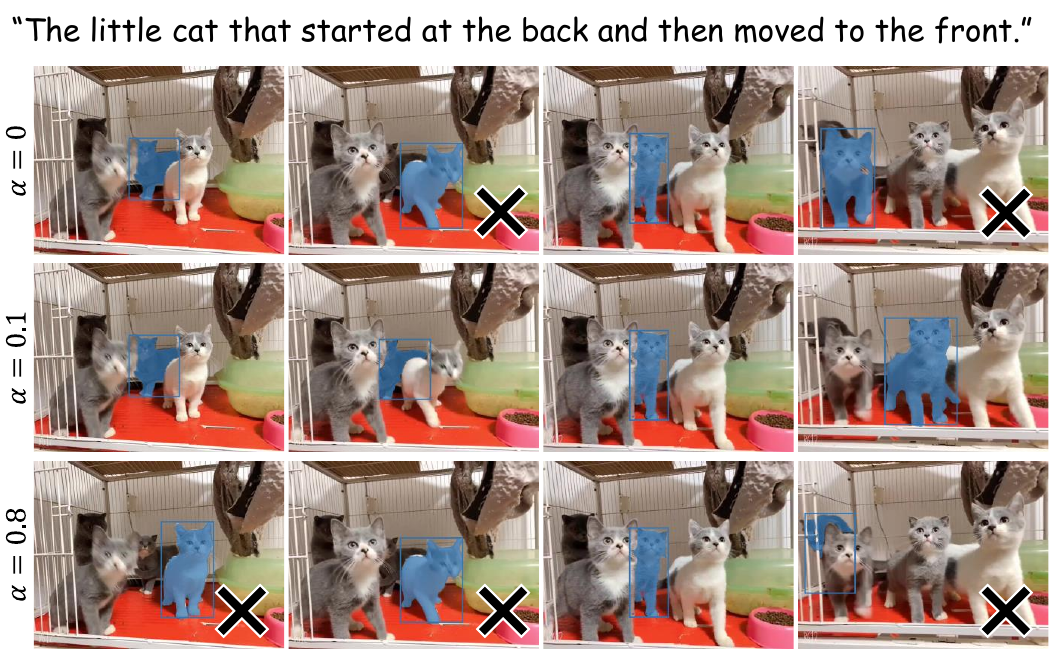}
    \caption{Qualitative impacts of $\alpha$ in memory-augmented tracker. We use \ding{53} to highlight the incorrect results.}
    \label{fig:alpha}
 \end{figure}

\textbf{Momentum Coefficient.} 
The momentum coefficient $\alpha$ in our tracker controls the amplitude of memory updating. We visualize a case in Figure~\ref{fig:alpha} to show its impact on temporal consistency. In this case, a smaller $\alpha$ (i.e., long-term memory) yields the best performance. This is because the identification in the initial frames is crucial as it captures the clue ``\textit{started at the back}''. 
While we set $\alpha=0.1$ by default in main experiments, it is promising to explore an $\alpha$-adaptive strategy in future work. 
In our main experiments, we set $\alpha=0.1$ by default.

\noindent

\section{More Visualizations.}\label{sup_visual}
\begin{figure}[b]
    \centering
    \includegraphics[scale=0.46]{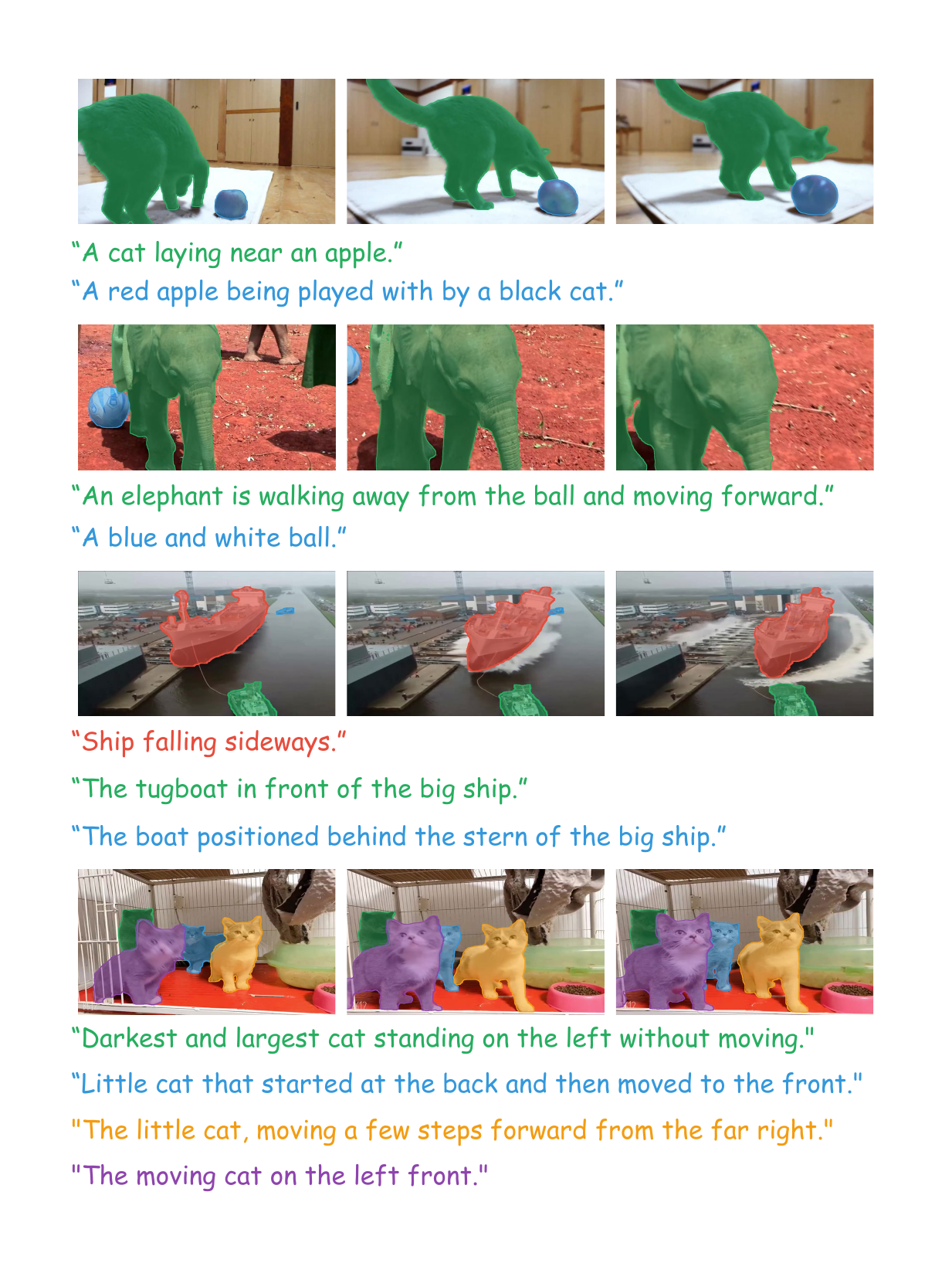}
    \caption{Visualization of our ReferDINO for multiple text references.}
    \label{fig:more_visual}
\end{figure}

We provide more visualizations of diverse objects in Figure~\ref{fig:more_visual} to demonstrate the robustness of our model.

\end{document}